# Taxonomic survey of resources for Hindi Language NLP systems

Nikita P. Desai and Prof.(Dr.) Vipul K. Dabhi

Dept of IT, DDU, Nadiad, Gujarat, India, 387001.

January 30, 2021

**Abstract**

Natural Language processing (NLP) represents the task of automatic handling of natural human language by machines.There is large spectrum of possible applications of NLP which help in automating tasks like translating text from one language to other, retrieving and summarizing data from very huge repositories, spam email filtering, identifying fake news in digital media, find sentiment and feedback of people, find political opinions and views of people on various government policies, provide effective medical assistance based on past history records of patient etc. Hindi is the official language of India with nearly 691 million users in India and 366 million in rest of world. At present, a number of government and private sector projects and researchers in India and abroad, are working towards developing NLP applications and resources for Indian languages. This survey gives a report of the resources and applications for Hindi language NLP.

## 1 Introduction

The field of Natural language processing can be formally defined as -

"A theoretically motivated range of computational techniques for analyzing and representing naturally occurring texts at one or more levels of linguistic analysis for the purpose of achieving human-like language processing for a range of tasks or applications"[69].
The naturally occurring text can be in written or spoken form.A wide array of domains contribute to NLP development like linguistics, computer science and psychology.The linguistics field helps to understand the formal structure of language while computer science domain helps to find efficient internal representations and data structures.The study of "Psychology" can be useful to understand the methodology used by humans for dealing with languages.
NLP can be considered to be having two distinct focus namely (1)Natural Language Generation(NLG) and (2)Natural Language Understanding(NLU). The NLG deals with planning to use the representation of language to decide what should be generated at each point in interaction, while NLU needs to analyze language and decide which is best way to represent it meaningfully.We, in this survey paper, concentrate on area of NLU for written text.Hence the NLP henceforth might be considered as NLU and vice versa.

**Motivation for designing Indian NLP systems** Hindi and English are the official languages in central government of India(GOI). Indian community faces a "Digital Divide" due to dominance of English as mode of communication in higher education, judiciary, corporate sector and Public administration at Central level whereas the government in states work in their respective regional languages [67].The expansion of Internet has inter-connected the socioeconomic environment of the world and redefined the concept of global culture.As per a report in 2017 by the companies kpmg and Google[1] 71% of users are not using digital applications as the contents in the applications were in English and not in local language. The report also claimed, in next few years, online digital applications would be highest growing category, followed by digital classifieds, digital payments, chat applications and digital entertainment. Thus, local language support is the need of hour at both national and international level.
Hindi belongs to "Indo-Aryan" category whereas English falls under "West Germanic" category of the IndoEuropean language family.Hindi has nearly ten thousand million speakers across globe [2]. In-spite of such wide usage, as per current statistics[3] 59.7% of websites use the English language text, while Hindi is used just 0.1% sites. Hence, developing Hindi NLU systems is hight. As per our findings, this effort is first of its kind covering the widest range of taxonomy for tools and resources. We also try to give comprehensive coverage of the aspects like open issues, challenges, current solutions proposed, etc. related to the NLP tools in Hindi language.

---

[1] https://bit.ly/kpmg_IL_internet
[2] https://bit.ly/native_speakers_IL
[3] https://bit.ly/website-languages

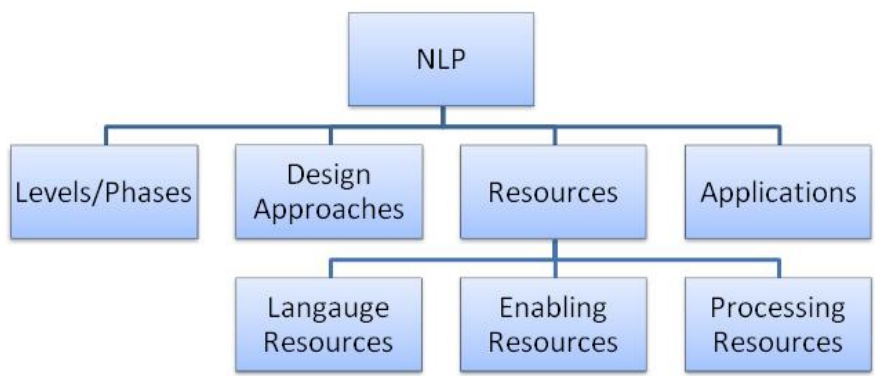

Figure 1: NLP taxonomy

**The taxonomy of NLP System** The taxonomy of NLP system with focus on "understanding" the "written" text is as shown in figure 1. As shown in figure, NLP system can be categorized based on various aspects like depth of text understanding, tools and resources used to build system, approaches used to develop systems and the applications which can use it.

The rest of paper is organized as follows.The various levels/phases of understanding at which NLP system can be developed would be discussed in detail in section 2.Next we would explore various approaches for designing NLP systems and the main evaluation methods typically used for judging the quality of any NLP system developed in section 3.The section 4 would discuss tools/resources and their current status in Hindi NLP systems.Next, in section5, we would present the survey of few major applications of NLP in Hindi.The section 6 would discuss the findings based on analysis of various aspects and ultimately we would conclude in section 7.

## 2 Levels of understanding

A NLU system must have considerable knowledge about the structure of the language including what the words are and how they combine into phrases and sentences to convey the meaning [69].The various stages to understand any text can be divided into phases. The phases arranged in hierarchy based on the complexity or depth at which they perform the processing are as follows-

1. Morphology: knowledge about smallest valid unit of text and what they convey is used here. This phase would help to understand the meaning of word using morphology rules.For example adding "ed" to action word indicates the associated action had occurred in past.Example- posted, ended, etc.

2. Lexical: Here system tries to associate the extra meaning to words which can help in understanding text. This is done by associating a part of speech(POS) tag to word like noun, verb, adjective etc.Example- "Ram ate apple" on tagging with proper noun, verb and inanimate noun respectively, clears the meaning about who is doer of act, what is done and what was object of the act.

3. Syntactic: Here the syntax structure helps to find the meaning of sentence.Generally the order and dependency of words in a sentence help us understand things like who did what to whom.For example: "cat killed mouse "and "mouse killed cat" differ only in syntax order yet they convey quite different meanings.To extract such knowledge we should use appropriate grammar rules for language, which can help us understand the roles of each word in sentence.

4. Semantic: In this phase disambiguation of word meaning is to be done, based on knowledge of senses of words.If a word has multiple meanings, in a sentence, it is disambiguated to give one correct sense/meaning. Example: "The river bank was muddy" has an ambiguity in meaning for "bank" which has many possible meanings including financial institute.But due to the presence of "river" it is given meaning of "land alongside water reservoir".

5. Discourse: This phase uses knowledge that helps to get context based understanding of entire text based on the connection between sentences. Example:-In sentence - "Ram is boy.He is smart", the pronoun "He" is understood to refer to "Ram" based on processing done in this phase, using method called "annaphora resolution".Another method which might be useful to understand the text, is by finding sentence usage in the text like if it is news article, finding the discourse components like main story, lead event, etc. might help to understand text better.

6. Pragmatic:This phase uses beliefs, goals in general like how humans reason, to understand the sentence. This phase needs world knowledge. And it tries to extract meaning which is not literally mentioned in text.Example:-Identify the book theme is "hatred" even though nowhere in text the words indicating hatred like "hate", "despise", "dislike" etc are mentioned.

As stated by Liddy et al [69], the levels can interact in a variety of orders and not strictly sequential.We can use the information obtained by higher level of processing to assist in a lower level of understanding.For example, the pragmatic level knowledge indicating the document you are currently reading is about a "share market", can help the lexical level to associate the "financial " sense, to the multi-sense word "bank". It is found that majority of naive NLP systems use only the lower levels, as these levels use units which are majorly governed by rules and hence easy to implement.The research in English NLP is gradually shifting from lexical understanding i.e.,using "bag-of-words" to compositional semantics which is using "bag-of-concepts" and next they are targeting for narrative-based models, which are based on pragmatics [28]. On the contrary Indian languages are still identified as low-resource and hence it is seen as challenge for the linguistic community to develop NLP systems for them[49].

## 3 Design approaches

For designing any NLP system, the basic steps are- appropriate data collection as per target application, analysis of data to build system, construction of system and finally testing of system with fine tuning to make it efficient system. For developing the NLP system, we have option of selecting any one of four major methodologies, as discussed below [69]-

1. Rule based approach or symbolic approach: Under this approach, the system is designed after deep manual analysis of language phenomenon. The appropriate symbolic mathematical schemes and algorithms are used to give best lexicons and rules for governing the NLP system. Few well known symbolic representation that can be considered are propositional logic, predicate logic, facts, rules, semantic networks etc. This is one of most basic primitive approach. The human expert needs to keep regularly updating the system to handle changes needed in ever growing NLP field.

2. Statistical approach:It uses various statistical machine learning models like Hidden Markov machine(HMM), Support vector machine(SVM), Conditional Random Field(CRF), Naive Bayes(NB) etc. These models take a large amount of tagged/annotated data (corpus) to statistically analyze and learn the language characteristics. And later the models can be used to understand any unseen text.

3. Connectionist approach:This approach is inspired by biology and psychology fields, which is based on knowledge of brain cells connection and its working to understand texts. Models based on this approach are called Artificial Neural Networks(ANN). The model can deal with unknown words/patterns, by adjusting the weight of connections. Recent advancement of this model is the deep learning model.

| Approach | Advantage | Drawback |
| --- | --- | --- |
| Symbolic | No need of large amount of data in designing system | Needs human expert else becomes brittle and non flexible when handling unseen inputs |
| Statistic | Automated design and updation | Needs large amount of good quality annotated data. |
| Connectionist | Adaptable for changes in real time data. | Needs huge amount of noise free observable data and lots of computation power to process it. |

Table 1: Design Approach summary

4. Hybrid approach-Now as shown in table 1, none of the design approaches is best for NLP systems and hence it is suggested to use a hybrid approach of combining them so as to design best possible system.

**Evaluation Methods**  To evaluate quality of any NLP system we would need some standard applicable methods. Following gives a informal explanation of few major evaluation measures generally used in NLP.

1. Accuracy - Ratio or fraction of correct results obtained by NLP system. It measures how close the system got to the actual result expected. The measure is majorly used in Text Classification(TC) like applications,where there are nearly equal amount of positive and negative instances in dataset .
The following three measures namely precision, recall and F1 score are widely used in situations where the source data is not balanced in terms of positive and negative samples.Such applications are the various information access based applications like Information Retrieval(IR), Information Extract(IE), Question Answering(QA) etc.

2. Precision-From all the data processed(retrieved), how much was actually correct(relevant). So it emphasizes to find the true positives.

3. Recall- From the entire data set how much relevant data was actually (retrieved) processed and how much was missed. Thus fetching all data from dataset might give high recall, but result in low precision and vice-versa.

4. F1 score- Gives a combined measure based on mean of precision and recall and thus helps to to give single evaluation of system.

5. BLEU(bilingual evaluation understudy) score -Helps to find how close the automatic NLP system is to a human's natural language processing.Used mainly in machine translation applications.

6. ROUGE(Recall-Oriented Understudy for Gisting Evaluation)-similar to BLEU, used in text summarization applications.

# 4 Resources/Tools

To facilitate building a complex NLP system for various utilities, some basic tools might be useful. The detailed taxonomy of such resources/tools is shown in figure 2.

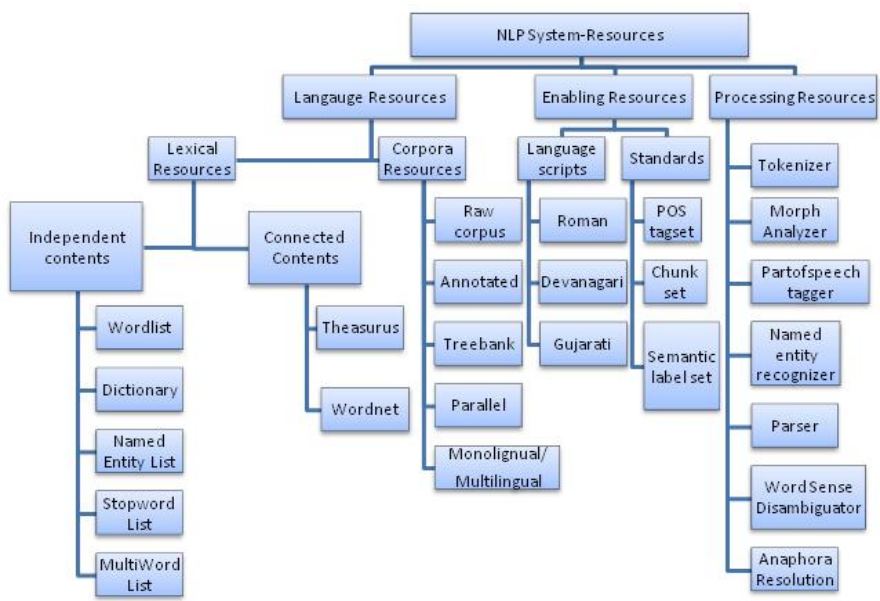

Figure 2: Tools/Resources in NLP

These resources would aid the NLP systems at various levels and in various applications. The typical characteristics of written text in IL which make designing of these tools challenging are as follows-

- Free word order- Typically English is fix order languages where the order of words in sentence indicates Subject(S) followed by verb(V) and finally the object(O). Example:- "Ram [S]gave book[V] to Sita[O]" which in Hindi can be translated to Hindi as "Raam ne(S) kitab di(V) Sita

ko(O)", but IL is not fixed order, and so the same sentence can also be written as "Raam ne Sita ko kitab di[SOV]", "Sita ko Raam ne kitab di[OSV]", "kitab Sita ko Raam ne di[VOS]", "kitab Raam ne Sita ko di[VSO]", "Sita ko kitab Raam ne di[OVS]".

- Spelling variations-Single term can have multiple correct spellings For example Hindi word अँग्रेज़ी :(angreji) meaning "English" can be spelled as अँग्रेज़ी, अँगरेज़ी, अन्ग्रेजि .

- Missing capitalization feature- In language like English, the proper nouns and first words in sentences start with capital letter. Example:-George, Washington, Microsoft etc. However IL does not have this distinction.

- Ambiguity between grammatical categories of words-Examples:- "Kamal" can be proper noun(person name) and common noun(lotus), "Pooja" is verb(worship) and proper noun(person name), "Ganga" is a person name and river name.

Following are few agencies actively working to provide these tools/resources.

1. TDIL:Indian Language Technology Proliferation and Deployment Center, Funded by Dept. of IT(DIT), Govt of India(GOI)[4]

2. IIT -B :Indian Institute of Technology-Bombay [5]

3. IIIT-H:International Institute of Information Technology, Hyderabad[6].

4. FIRE:Forum of Information Retrieval, held under IRSI (Information Retrieval Society of India) [7]

5. LDC-IL:Linguistic Data Consortium for Indian Languages, managed by CIIL(Central Institute of Indian Languages), Mysore. Funded by MHRD, GOI[8].The LDC is a open consortium of universities, libraries, corporations and government research laboratories which is hosted at world level by the University of Pennsylvania.

6. CDAC: Centre for Development of Advanced Computing (C-DAC) is the premier Research and Development organization of the Ministry of Electronics and Information Technology (MeitY) [9].

The various resources which are needed to work in coordination for efficient working of any NLU system can be broadly classified into 3 groups. The categorization of tools is based on their goals. The tools with goal to provide processing modules are called processor resources, the resources which provide standards to enable consistent representation are grouped under enabling resource and the resources which provide language specific data sets are named language resources.A typical NLP system would need usage of tools from each of these categories.

**4.1 Language Resources**

Every NLP system would need some input data for analysis, which it can further utilize in designing the appropriate automated system.This resource is providing requisite language specific data to system, hence it is named as "language" resource. They can further be classified into two categories, lexical and corpora.The lexical resource gives data in smaller units like word.Whereas the corpora is a collection of big texts spanning multiple sentences.

**4.1.1 Lexical resources**

These are the resources which give information at the smallest possible unit of text, called lexemes. A system can understand the data at lexical level using these resources. The resource can be further classified, based on existence of connections between elements in the resource as -

1. Independent contents :
   (a) Wordlist or vocabulary- list of all words in a data collection(corpus). Example:- "Rose is red" has the vocabulary of { "red", "rose", "is"}.

---

[4] http://tdil.meity.gov.in/
[5] http://www.cfilt.iitb.ac.in/
[6] https://www.iiit.ac.in/research/centres/
[7] http://fire.irsi.res.in/
[8] http://ldcil.org
[9] http://www.cdac.in/

(b) Multiword expression list(MWE)-List of group of words which combine to give one meaning. They can be idioms, compound nouns(N) and verb(V) constructions[112]. Examples:-વટ પડી ગયો (vaat paadi gayoo) roughly meaning you are impressive, બાગ બાગીચો (baaga bagiichoo, garden) (N+N), કામ આવવું ( kaama av vu, to be useful) (N+V), દોડી જવુ (doudi javu, to rush )(V+ V ).

(c) Dictionary -Gives root word plus grammatical information of word like Part of speech(POS) or translated /transliterated language pair of words between languages, usage examples, etc. Ex:-આશા -transileration "asha", translation -hope(in English), categories :Noun and Verb.

(d) Named Entities list (NE)- An entity is an object or set of objects in the world. Examples: India is location entity, Ram is person entity, 12:05 am is time entity etc.

(e) Stop word list -It is list of words not having any significant meaning.

The list of standard resources with other details, available for IL is given in table 2. As is observed we have considerable resources of this category for Hindi languages, released by Government and individual agencies. Besides the regularized resources listed in table, there are few other well designed resources directly shared by researchers like a stop lemma list of 311 words in Hindi provided by Venugopal Gayatri[131] [10].

Table 2: Language Resources : Unrelated lexemes

| Provider details(year) | Quantity | Details |
| --- | --- | --- |
| TDIL-AnglaMT:CDAC, Noida (2017) | 51K-Lexeme, 14-MWE | Hindi data in Txt format |
| TDIL-EILMT:CDAC, Pune (2016) | 14K-health and 39 k-tourism | Dictionary of Eng-Hindi pairs |
| TDIL-Sandhan: IITB (2016) | 2k-NE and 624 -phrases | Tourism data for Hindi in Unicode |
| TDIL-Sandhan: IITB (2016) | 35K-translation and 34K-transliteration | Tourism based dictionary of Hindi-Eng in Unicode |
| FIRE: IIT K (2013): | 117k-wordlist(with frequency), 30k(transliteration) | For Hindi in Unicode -UTF8 |

2. Connected contents(with relations):

   (a) Thesaurus-Similar words are grouped under same head word. Each entry consists of a list of comma separated values, with the first term being the root word, and all following words being related terms. Ex:- "hope", "faith", "optimism" etc.. For English there are few well defined thesaurus available in public domain like Roget, Moby etc.But for Hindi no such well designed comprehensive resources are available as of now.

   (b) Word net(WN) - They are composed of synsets(set of synonyms) and semantic relations amongst them. The data is represented as a graph with lexical or semantic relations as arcs between nodes of synsets. The semantic relations are between synsets and lexical relations are between words. Examples of few of relations in Hindi Word Net(HWN) are described in table 3

| Relation | Example |
| --- | --- |
| Hyponymy and Hypernymy (is a kind of) | कार (car::hyponymy)->वाहन (vahn:automobile ::hypernymy) |
| Meronymy and Holonymy (Part-whole relation) | जड़ (jaR; root:: merorymy ) -> पेड़ (peR ; tree::holognymy), |
| Semantic relation Between verbs Entailment | खर्राटा (kharraaTaa lenaa; snore-> सोना (sonaa; sleep) |
| Troponymy(verb elaboration) | मस्कQना muskuraanaa; smile-> हंसना (hansnaa, laugh) |
| Antonym(lexical relation) | मोटा (moTaa; fat) -> पतला (patlaa, thin) |
| Causative(lexical relation) | खिलाना (khilaanaa; to make someone to eat) -> खाना (khaanaa ; eat) " |

Table 3: Examples:HindiWordnet relations

The manually created Hindi wordnet design is inspired by English wordnet which is maintained by Princeton University.It contains four categories of words namely noun, verb, adjective and adverb.Each entry in the Hindi word net has synset, gloss containing example and definition of concept and position in ontology.An example taken using visualizer developed for Indowordnet [30] is shown in figure 3.

---

[10]https://github.com/ gayatrivenugopal

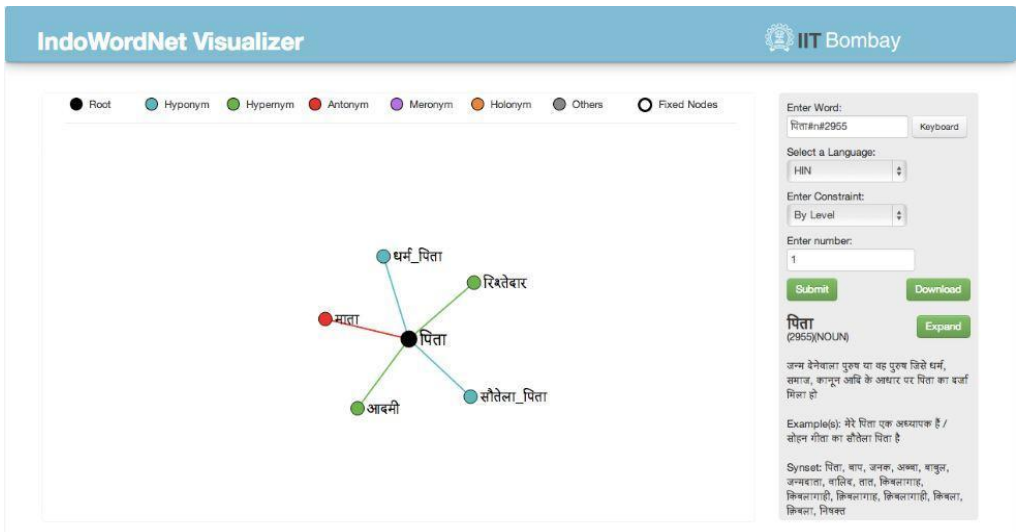

Figure 3: IndoWordnet sub-graph for Hindi word "pita"(father)

Basically, there are two approaches for generation of WordNet. First, merge model where an existing thesaurus is manually transformed into a wordnet. Hindi wordnet can be said to have used this first approach. Second is Expansion model which utilises an already existing WordNet template to build a new wordnet. In this approach, all senses listed in source WordNet are translated to target WN and all relations among senses of original WordNet are preserved in the generated WordNet[51]. IndoWordnet is a linked structure of word nets of major Indian languages, including Hindi which is based on the expansion from HWN. [25][23].Till April 2020,And HWN has 40465 synsets for 105423 words. The Indowordnet are available to download [11]. The English WN has 117659 synset till April 2020, which indicates the HN wordnets are still having lots of scope of enhancement. The major hindrance in IL fully automated wordnet generation is polysemy and homonymy, which needs sense selection to be done manually.

Few other related resources are also designed like Indowordnet linked with SUMO(Suggested Upper Merged Ontology). SUMO is language "independent" hierarchical structure of concepts which is free and publicly available whereas Wordnet is language specific. For example consider there are two statements s1=उसे गाना पस्नद है and s2=उसे संगीत पस्नद है , where meaning of s1 is " she likes singing" and s2 is "she likes music". Then using Hindi wordnet it is not directly possible to infer s2 from s1, as there is no hypernymy-hyponymy relation between singing and music.But if SUMO used, there is a relation which connects this concept.22148 Hindi synsets are connected to SUMO by Bhatt et al [22].Such linking of a language independent semantics of SUMO with Hindi word net can be useful for many NLP text applications. Polarity lexicons are also a noteworthy resource where entries are tagged with their prior out of context polarity based on the way humans perceive them using their cognitive knowledge. A system where the scores are copied from English sentiwordnet (SWN) based on their matching Hindi word found in bilingual dictionaries(Shabdkosh and shabdanjali), is suggested by Das et al[36]. They also used the Hindi wordnet based expansion approach to add more entries based on the relation of synonym and antonym. It is semiautomatic method with human intervention to validate system. Total of 22, 708 entries for Hindi are reported, however it is not tested on any application.Another polarity lexicon with 3 polarity positive, negative and objective on adjectives and adverbs, is designed by [10]. They used HWN and seed data to do graph based expansion using breadth first search using synonymy and antonymy relations to assign score. It gave 79% accuracy when applied on sentiment classification of Amazon product reviews translated to Hindi. And they also claim 70.4% human annotator agreement of lexicon.

#### 4.1.2 Corpora

A text corpus or corpora is a large set of texts. The text directly obtained from any source without adding any extra information is called "raw text". The text when added with extra information is called annotated text. When given with such tagged dataset, the system designer can analyze and find patterns/rules /data structure to which can be used to develop the system. The various varieties of annotated text based on types information added are listed below-

---

[11]http://www.cfilt.iitb.ac.in/wordnet/webhwn/downloadIWN.php

1. POS tagged- Partofspeech of each word in sentence in the text is tagged.

2. Morpheme tagged-The words are tagged with reference to their constituent smallest units(morphemes).

3. Parse tree tagged-Dependency or constituency parse structure information added to sentences. Also called "tree banks". It can be useful at the syntax level design.

4. Co-reference tagged- Tagging for resolving pronouns in related sentences like "Ram is a boy.He is in second class". It can help to do discourse analysis of text.

5. Mono or bi or multilingual -The text is written using single, two or multiple languages. Corpus having multiple languages in same text might be useful in applications like multi language IR, cross language IR etc.

6. Parallel-Giving matching annotations on sentence pairs in two different languages. It might be useful to design MT systems.

7. Comparable-Gives the text under similar circumstances like electronic gadget manual in more than one languages.

8. Domain specific or general-Text identified to be on specific domain like movie review, health, agriculture or general text.

The table 4 gives the general corpora resources available currently for Hindi which can be used in any application.

Table 4: Language Resources - Corpora

| Provider details | Size (sentences) | Domain and Type | Other details - Language, format, etc |
|---|---|---|---|
| FIRE (2012-13) | words-30K | general: Morphology annotated | Hindi. |
| TDIL-ILCI (2017) | 25k, 25k, 38k, 37k | health, Tourism, Agriculture and entertainment:Parallel corpus | Hindi- English pairs, UTF-8, BIS Tagset. |
| TDIL-ILCI (2017) | 36k(H), 30k(G), 31k(E) | General: POS tagged and chunked | Hindi and English, UTF-8, IIIT-H Tagset. |
| TDIL-ILCI (2017) | 11k, 4k, 14k for Guj and 39k, 3k, 14k for Hindi | Toursim, Agriculture and Health: Parallel corpus | English to Hindi pairs, XLS format. |
| IIIT -H (2016) | 3k | general: Treebank(phrase and dependency) | Hindi ; SSF (Shakti Standard format). |
| IIIT- H (2018) | 3600 | general: co-reference annotation. | Hindi |
| EMILLE [12] | 12 M words | Monolingual | for Hindi. |
| EMILLE | 2L words | Parallel | Eng to Hindi. |
| UFAL, Hindiencorp [13] | 44M sentences (787M words) | Monolingual. | Hindi |
| UFAL, Hindiencorp | 273 k sentences (3.8M words) | parallel corpus | Eng-Hindi |
| LDCIL | 28L words, 1M words | Domain specific:Raw | Hindi text;domain specific;Unicode, XML storage. |

### 4.2 Processing Resources

The processing resources work like small building blocks or modules to "process" data and combine to make a complete natural language understanding system. In general two major types of approaches are observed to design such systems- hand crafted Rule based and automatic machine learning based.

#### 4.2.1 Tokenizer

The original text is a sequence of characters. Before any syntactic analysis of the corpus can commence, two tasks are needed to be done by tokenizer tool. First task is sentence identification by using techniques like sentence boundary detection (SBD) and another is word identification from given stream of characters. Word tokenization is done based on white spaces in Hindi. The word tokenization identifies two types of tokens: simple and complex.The simple type is corresponding to units whose character structure is easily recognizable, such as punctuation, numbers, dates, etc.While the second type of token will undergo a morphological analysis to identify its constituents. Few examples of complex tokens are "uncomfortable", "unhappiness" etc[46]. There are issues which make SBD tricky like the multipurpose usage of punctuation marks especially dot .e.g., 'Ms.', 'U.A.E'; 'http://bit.ly/'; '1019.0', etc.Generally the punctuation marks like ".", exclamation mark "!", and question marks " "" indicate sentence boundaries in English texts. For Hindi same set of symbols are used,with exception of a special delimiter symbol for sentence boundary "|" , instead of ".".Once the tokens are extracted from the input text, it needs to be assigned a unique unambiguous category like date, roman numeral etc.For example, '1824' could be of the type 'Year' or type of 'cardinal number'. The research in Hindi tokenization is majorly focused on social media text written in roman script having informal flow and using mixed Hindi and English(Hi-En) language sentences. Typical issues in such texts are -

- Sentences written in multiple languages without any marking while switching, so difficult to set up boundary between sentences. Example:- "admin plz post kar dena 5th time likh rha hu...next month sept. me mera exam hai".Here there are two separate sentences "admin plz post kar dena 5th time likh rha hu" and "next month sept. me mera exam hai"

- Ambiguity in language identification. Example:- "Mom to man gai par papa katai man ne ko taiyaar na the" Here "man" and "the" should not be identified as English as they are used as Hindi words

- Using semi-broken words. Example:-sept., bro, sis etc

Research was done on SBD by Rudrapal et al [99] in 2015, on codemixed Hi-En data having 554 tweets. The tweets were manually divided into 1225 sentences by two human annotators. Researchers had used two approaches rule based system and statistical. Under statistical approach they used 3 machine learning models -Naive Bayes(NB), Conditional Random field(CRF) and Sequential Minimal Optimization(SMO). The F-measure results were given to be 64.3% by rules, 84.5% by NB, 84.3% by SMO, and 56.9% by CRF.The authors analyzed that when code mixing occurs at word level like in "100ka", "girl-sachhi" etc, SBD fails.Another rule based approach for SBD on Hindi-English codemixed data with accuracy of 78.58% is reported by Singh et al[110] in 2018 on self extracted gold corpus of 16K sentences.The system failed when the sentences were not following any patterns given in rules.

**Text normalization**  In Hindi text, it is found that many words are mixed with English words making it a bilingual text. Further numbers could be represented by Arabic or native numerals, in the same text.So it is essential to do text "normalization" once the token is identified. Text normalization involves tokenization of input text into tokens, identification of Non Standard Words (NSW) and representing the NSW in standard representations. It is found by Goyal et al[45] that Hindi has maximum 3 spelling variations in a text of 1 Lakh words, which can be handled by replacing all variations to standard form, based on rules.

Panchapagesan et al[83] have given method for text normalization and disambiguation in 2004, for Hindi text. The authors used manually identified regular expressions to find the tokens and in next phase, decision learning based statistical method was used for disambiguation. EMILLE corpus was used for training the decision tree classifier.The accuracy of float/time of disambiguation is 98.6% with just one surrounding token taken for learning (context window size), whereas for number/year the accuracy of 96.1% can be achieved with context window of 3 size.

### 4.2.2 Morphological Analyzer(MA)

Morphology analyzer's goal is the study of inner structure of words(lexemes). A word contains morphemes. A morpheme is the smallest meaningful morphological unit in a language. A word can have two constituent morphemes namely root or base and affix. The "base" morpheme contains major component of word's meaning. It has a lexical category like noun, verb, adjective, adverb and can be used independently without using any other morpheme. The "base" is the head word or "lemma" of various set of words. The "root" morpheme are those basic form of word that can't be used alone without affixes. The root word are part of base words.However, in literature of MA, the root and base words are used interchangeably.

A prefix, suffix or infix is a meaningful affix that doesn't function as a word on its own but can be attached to a root or base word. Prefixes (like pre-, anti- or de-) go at the beginning of a word, while suffixes (like -tion, -ness and -ment) go at the end and infixes (like -s) are used in middle of word. Examples:- word with suffix "rough + ness"; Word with prefix "un+grateful"; word with infix "passer+s+by"; distinct base words - "audio, audience and auditorium" having same root word "aud".

Inflectional morphemes and derivational morphemes are two main types of morphemes. An inflectional morpheme is a suffix that's added to a word to assign/change its grammatical property such as its number(singular/ plural), mood(verb form like interrogative/ fact/ wish/ command etc), aspect(progressive/ perfect), tense(past/ present/ future etc), person(first/ second/ third), animacy (living /inanimate), voice(active/passive) or case(direct/oblique). However, an inflectional morphology can never change the grammatical category of a word. Example:-For the word "look" the various inflected words formations are "looking"(progressive aspect ), "looked"(past tense, perfect aspect) and "looks"(present tense). Second type of morpheme is derivational where affix is added to a word in order to create a new word or a new form of a word. A derivational morpheme can either change the meaning or the grammatical category of the word. Examples :Pure →Impure (Change in Meaning), Help (verb) → Helper (noun)(Change in Grammatical Category).In Hindi :म+ मेरा = ममेरा (Pronoun to Adjective )

Indian languages are considered richer in inflectional morphology as compared to English, hence making it more challenging. IL have more average number of inflections for words then in English. For example, the English noun word "boy" has only two inflections, giving words "boys" and "boyhood". Whereas it's counterpart Hindi word लड़का: (ladkaa) has following inflected words लड़के for- plural in direct case, लड़को for plural in oblique use, लड़की meaning girl, लड़कपन meaning boyhood.

**Methods to find morphemes-Stemming and Lemmatization** The term "conflation" or "stemming" is used to denote the act of mapping variants of a word to a single term or "stem". Stemming finds the stem whereas lemmatization removes inflectional endings and returns the base or dictionary form of a word. Stemmers are used in text retrieval systems to reduce index size. Reducing morphological variants to same stem eliminates the need of using all variants for indexing purpose. This helps in improving recall. However, it may sometimes degrade performance[84]. Based on working behaviour, stemmers can be categorized as lightweight or iterative. The lightweight finds affix in just one iteration in contrast to iterative methodology. First try for Hindi stemmer generation was done in 2003 by Ramanathan et al[96]. It was rule based stemmer for only inflected words which failed to give correct stems for 17% words from total of 36k words. Unsupervised stemmer for Hindi was suggested by Pandey et al[84], which gave accuracy of 89.97% which when compared to lightweight stemmer in same setup was 19% better. However the authors did not apply normalization to handle word spelling variations when designing the stemmer. In 2013, rule based lemmatizer was designed by Paul et al[90], which gave 89% accuracy in Hindi, on self made corpus of 2500 words, for which the base words were made using paradigm approach. In 2014, a comparison between stemming and lemmatization by Balakrishnan [11], found that lemmatization outperformed stemming by 13% in task of information retrieval, as lemmatization helps in more detailed analysis as compared to stemmer.

**Approaches to do morpheme analysis** The goal of analysis of morphemes found is give information about the word's grammatical details like gender, animate, tense, aspect etc.

1. Using "exhaustive lexicon" of all possible word forms. For example if word is "ladkaa(boy)" store all its forms like "ladkey (boys)", "ladako (boys)" with details like plural direct case, plural oblique case respectively for each of the form. The main drawback of this technique is huge storage requirement.

2. Using "paradigm approach", where for all words following same patterns of word formation, make single entry in paradigm table. For example in Hindi, word formations of the words "laDkaa" and kapaDaa (cloth) are same so they both can be represented under same name "laDkaa" in paradigm table.Major issue with this method is the need of language expert to identify the paradigms.

In 2005, Shrivastava et al had developed the rule based inflectional stemmer and paradigm based morphology analyses tool[108].The stemmer was using modified rules also to recreate the root after identification of stem.Thus stemmer gives the root, suffix and grammatical category and the morphological analyzer tool provided the detailed analysis like gender, person etc.The stemmer gave 100% precision when tested on 12k words. In 2008, Goyal et al[44] made a morphological analysis and generator GUI tool for Hindi Language using paradigm approach and elaborated mainly on the database design in their reported paper, without any details on evaluation results of tool.In 2011, a method for morphological analysis for verbal inflections in Hindi was given by Singh et al [119]. The authors used hand crafted rules to find the root word and then using lexicon of 1500 verbs, the grammatical information was extracted from lexicon to give the detailed analysis of given word.The manual verification of system on self made corpus of 13160 verb forms was claimed to give 99% accuracy. In paradigm based approach for Hindi, when the derivational morphemes were also identified along with inflectional by Kanuparthi et al[59], improvement of 5% was observed as compared to system using only inflectional morpheme identification.

Statistical methods using linear SVM(support vector machine) were tried by Malladi et al[72] for Hindi and they reported accuracy of 82.03% .In 2015 research reported by Soricut et al[124], unsupervised learning using word embeddings on morphological rich languages like Uzbek, it was found to that MA gave accuracy as high as 85%.However the research did not use any IL for testing method.

For Hindi language a tool for morphological analysis is released by IIIT-H, which is rule based and claimed to be having 88% coverage.
The table 5 gives the summarized status of research in MA.

| Challenges | Solutions papers | Research gap |
|---|---|---|
| Find stem of highly inflected words. | [96] , [84],[126] [4][107][38] [86] | Spelling normalization ,codemixing not considered. |
| Find dictionary form of word | [90] [108] | Rule based ,testing not standardized. |
| Find paradigms for analysis | [108], [102] | Testing done on limited categories of words. |
| Derivational morphemes analysis | [126] | Not comparable accuracy as inflected analysis |
| Find rules for analysis | [119],[9] | Verbs only considered;Result not at par paradigm based. |

Table 5: Morphological Analyzer Research: Analysis and Findings

| Challenges | Solutions papers | Research gap |
|---|---|---|
| Ambiguous tags. | [42] [19] [35][109] [88] [118] [80] | Different tagsets used by all. Mostly statistic methods used. Self tagged corpus for training /testing. |
| Code mixed words. | [54]. | only tested on hindi-English mixed data. |

Table 6: POS tagger research: Analysis and Findings

### 4.2.3 Part of speech(POS) tagger

The POS Tagger is a piece of software that reads text and assigns appropriate syntactic categories called parts of speech to each word.Examples of POS categories-noun, verb, adjective, etc.The major challenge for POS tagger is identification of the best tag for a given word in case when the word has multiple possible tags. For example in Hindi:गंगा ने गंगा में डुबकी लगाई ! (Ganga ney ganga mey dubki lagai;Ganga took a dip in Ganga), the first गंगा :Ganga is referring a human entity while second is referring the river, hence they should be tagged proper noun and common noun respectively.To do this disambiguation of tags, the taggers need rules which can be hand coded, or learnt by machine from POS annotated corpus. Another rule generation approach is given by Brill[27], where the rules are learnt automatically by system based on concept of transformation based learning(TBL). Here, initially the most frequent tag is assigned to each word found by corpus analysis or from lexical resource like dictionary. In case the annotated corpus is used it falls under supervised technique and if no tagged corpus is used and task is done manually it becomes unsupervised technique.In next step, assigned tags are verified manually or with actual annotated corpus. And based on errors found, correction rules are automatically formed. This method gave accuracy of 95% on tagging for English corpus.

During a contest held by IIIT-H in 2007[19] for Hindi POS tagging and chunking, it was found that hybrid method of using CRF followed by transformation rules to correct any errors, gave accuracy of 78.66% which was best among HMM, Decision Forest, and basic CRF. This contest used 24 POS tags, annotated corpus of 20k and test corpus of 5k. A survey on taggers for IL done by Kumar et al[63] and Antony et al[7], found morph analysis is very important for improving accuracy of tagger. Another observation was missing common accepted tagset for IL and no participants used rule based tagger.Also all methods for Hindi were mainly on statistical or hybrid approaches. The best result 93.45% accuracy was obtained in the approach which had exhaustive morphological analysis data and decision tree based algorithm[118].Max entropy model using stemmer, morphological analysis from lexicon and 15562 words manually tagged with 27 different POS tags gave 94.38 when designed by Dalal et al% [35].It was proved by Shrivastava et al[109], using only simple lightweight stemmer can give considerable accuracy of 93.12% with hidden markov model using 81 k tokens tagged with ILMT tagset. In 2012 Garg et al[42] designed a hand crafted rule based tagger using 30 tags and tested on 26k words and achieved 87.55% accuracy.The tokenized word are first searched in the lexicon to find category.Later using rules the untagged words are given POS tags based on tags of neighbour words. In-case the surrounding words are also not given any tag, no rules are applicable and system fails to give tag to current word. The connectionist approach using artificial neural network, trained with 2600 sentences having 11500 words was used by Narayan et al in 2014[80] and result of 91.30% was reported on selfmade tagset of 27 tags. In 2015, Jamati et al[54] worked on codemixed English-Hindi social media text. They used customized tagset using universal tagset given by Google, Indian tagset given by TDIL, and twitter specific tagset. Total 1106 messages were manually collected and tagged. They applied statistical models CRF, naive bayes(NB), Random Forest(RF) and sequential minimal optimization(SMO) and concluded RF was best with 73.3% accuracy.However the CRF was only marginally poor compared to RF. If better annotated corpus with all possible word utterances are included in training, the accuracy can be further enhanced.

Currently POS tagger tool, for Hindi is released by IIT B[14] which is using CRF model. The table 6 gives summary of research efforts for POS tagger.

---
[14]www.cflit.iitb.ac.in/tools.html

| Challenges | Solutions papers | Research gap |
|---|---|---|
| Ambiguous categories | [87] [60][111][34] [94, 19] | Testing done on self made corpus and/or tagsets; Language specific rules used for error corrections; fail for noun categories . |
| code mixed language | [103] | Not comparable accuracy as monolinguistic chunker .Used small chunkset. |

Table 7: Chunker : Analysis and Findings

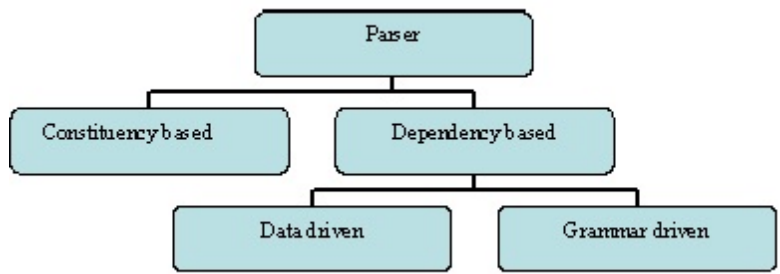

Figure 4: Parser Types

### 4.2.4 Chunker

Shallow parsing or light parsing or chunking is the analysis of sentence to identify its constituents or "phrases" at a shallow level without much emphasis on internal structure. It groups the words into phrases, based on the linguistic properties(POS). Example-A noun modifier should be associated with noun and so on.The phrases or "chunks" are needed to be minimal and non overlapping.Chunker shows the relation amongst the chunks in a sentence.

Chunker can be used by applications like Question Answering and Information Retrieval, which don't need a detailed analysis of "each" words connection to a sentence.For example, to retrieve any information, even if the overall sentence structure is not proper ,but at-least the phrase is correct ,some rough meaning can be associated and hence extraction can be done. Chunker can also be used as a step for making a parser, wherein once the overall structure is determined ,in depth analysis needed by parser can be executed. There are separate chunk tag sets related to chunk boundaries like B(Beginning), I(internal),which indicates word positions with reference to chunk and another tagset to indicate names of chunk like NP(Noun phrase), VP(Verb phrase) etc. Example:The sentence in Hindi सीता(Seeta) ने (ne) एक(ek) केला (kela) खाया (khaya), has the chunks NP-Seeta ne, NP-ek kela, VP-khaya. Further the words "Seeta" has the B-tag and "ne" has I-tag, similarly each first word in the phrase has B-tag and each second or internal word has the I-tag.

The first attempt for Hindi chunker was made in 2005 by Singh et al[111], who got accuracy of 91.70%, using HMM which used 2 Lakh words annotated by POS and chunk labels, provided by IIIT-H.However they used only 2 boundary tags, 5 chunk labels and the Penn tagset for POS tagging.Next year researchers Dalal et al [34] claimed the chunker efficiency depends on POS tagging accuracy and designed max entropy model to get 82.4% accuracy on data given under a contest on ML, with 35000 words having 26 POS tags and 6 chunk tags.In a contest held in 2007[94, 19], when same dataset was applied for TBL, CRF, Max Entropy, Decision Forest, Rule based and HMM +CRF approach, the best was HMM + CRF with 80.97% accuracy. Chunker for Hindi-English code mixed data was attempted in 2016 by Sharma et al [103].They used CRF to get accuracy of 78%.Eight bilingual speakers annotated 858 sentences using Universal POS tagset and self designed chunk tagset.The designed tool is released for public use [15].Another tool based on CRF with TBL is released by IIIT-H[16] which has accuracy of 87% but it is only for Hindi.

The 7 gives a concise summarization of research efforts for development of chunker tool.

### 4.2.5 Parser

Parsing is a process of taking a sentence and assigning it a suitable structure which can help NLP system understand relationship amongst word in sentence.Parsing process needs two components - parser(procedural/processing component) and grammar(declarative component).The figure 4 shows the types of parsers.

The grammar is specific to a language, while parser can be designed to work on any language.However both depend on "formalism" or "inter relations" to be used in parsing[16].A grammar can be constructed by hand by a linguist, and can also be induced automatically from a

---
[15]bit.ly/csmt-parser-api
[16]http://ltrc.iiiit.ac.in/analyzer/hindi

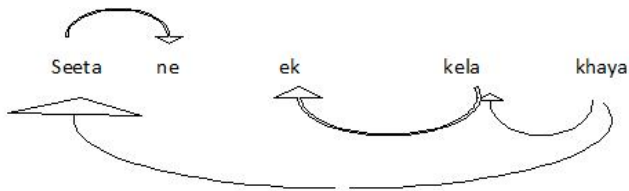

Figure 6: Dependency parse tree

"treebank" (a text corpus in which each sentence is annotated with its syntactic structure). Languages which have fixed word order sentences, can use the context free grammar or constituency based grammar formalism.Whereas for IL, which are having comparatively free word order, the usage of dependency grammar is more suitable which is supported by "Paninian formalism". The type of syntactic analysis a sentence gets depends on the type of grammar used. Constituent-based grammars focus on the hierarchical phrase structure of a sentence.The words belong to lexical categories(Noun/Verb etc) and they group together based on "phrase structure rules" to form phrases(NP-noun phrase/VP-Verb phrase etc) which in turn make sentence.The phrase structure rules define the correct orders for constituents in sentence. Another mechanism for syntactic analysis is using dependency analysis which use dependency-based grammars.This grammar focuses on relations between words.There are mainly two types of words in sentence, dependent(child) and parent(head) word.The directed relation(label) is used to indicate the relation from dependent to parent word.Examples for the parse trees under both mechanisms are shown in figure-5 and 6.The Hindi sentence used is "Seeta ne kela khaya" meaning "Seeta ate a banana".The tags used in the constituency tree are - NNP-Proper Noun , PSP-Postposition, VGF-finite verb chunk , VM-Main verb, QC-cardinal number, NN-Noun.The tags are used from IIIT-H tagset.

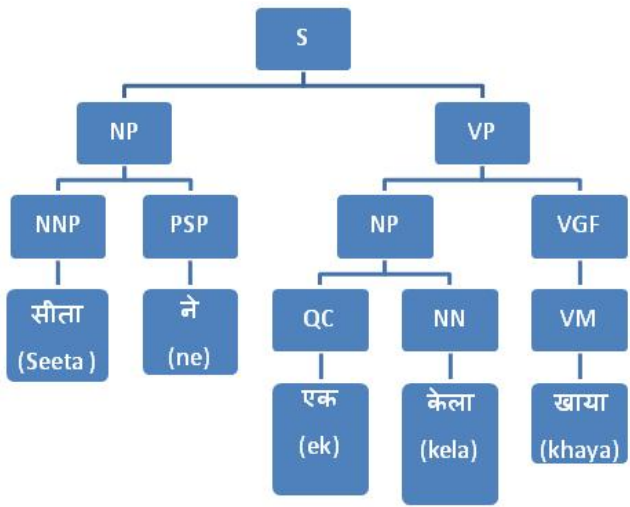

Figure 5: Constituency Parse tree

The dependency driven parsers can be further classified as data driven or grammar driven.The grammar driven parser view the parsing as constraint satisfaction problem where the parses are eliminated which do not satisfy given set of constraints.Data driven parsers use corpus for finding probabilistic model of parsing. The metrics used to evaluate dependency parsing results are Unlabeled Attachment Score (UAS), Label Accuracy (LA), and Labeled Attachment Score (LAS).UAS is the percentage of words in the sentences across the entire test data that have correct parents. LA is the percentage of words with correct dependency label, while LAS is the percentage of words with correct parent and correct dependency label. The few major issues which need to be handled when designing dependency parser for Indian languages are non-projectivity, long distance dependencies, complex linguistic phenomena, and small annotated corpus sizes. For Hindi, in 2009, a rule based parser based on POS tags, was designed by Bharati et al[17].The approach used rules designed after analysis of 2100 sentences in a treebank(HyDT-Hyderabad Dependency treebank), to give average precision of 79.1% for LA and 71.2% LAS when tested on 300 sentences for inter chunk relations.Whereas for intra-chunk (labels within a chunk of words) 96.2% of LA and 90.4% of LAS is reported after testing on 200 sentences.Another research by Ambati et al[3] in 2010 proved MaltParser[17] which is shift reduce parser was better than graph based MST parser[18] for Hindi. They also proved that features provided by chunker and some morphological features help improve the parser accuracy. However morph features of gender, number and person features did not help.The achieved results were 91.8% UAS, 83.3% LAS and 85.3% LS with MALT. The parsing was done in two stage, first intrachunk and then interchunk on the chunk heads found in first stage. Bharati et al [20]had first proposed a constraint based parser design for IL in 1990, which was based on relations used in Paninian grammar called "karaka" relations. Subsequently in 2009, a two stage dependency constraint based parser was suggested by Bharati et al[18].In first stage the parser tries to capture all verb argument relations like karma, karta etc and in second stage it marks the compounds relations like conjuncts.It was found that this method gave 76.9% accuracy for LA and 75% LAS which was better then data driven parsers MST and Malt on same test conditions.In a contest held by IIIT-H in 2010[50], the annotated corpus of nearly 3000 Hindi dependency sentences using 59 fine grained syntactic-semantic labels and 37 coarse grained labels were given to participants.It was found the best results were obtained on coarse grained labeling.The winners had used Maltparser to give 90.91% accuracy for LS.Another transition based shift reduce parser gave best UAS of 94.78% and LAS of 88.98%. In 2011 Begum et al[15] used the MSTparser to do parsing using verb conjunct features along with other features.They reported LAS of 69.05%, UAS of 85.68% and 72.29% LA.They used the datasets from HyDT treebank.Semantic category of word was taken from hindi wordnet . In another experiment Kukkadapu[62], used 3 data parser tools Maltparser, MST parser and Turbo parser[19] for Hindi parsing.They reported result of 96.50% UAS, 92.90% LA, 91.49% LAS using voting method on data with gold POS tags.Also the Turbo parser was found to best among the three.The system failed for projective sentences.

---
[17] http://www.maltparser.org/
[18] https://sourceforge.net/projects/mstparser/

Based on the analysis of work done in parser, it can be concluded that following are still open research issues -

1. Handling of non-projective sentences. Example sentence:- "Ram ney jaisa kaha maine vaisa kiya" ,where the link between "vaisa" to "kaha" is making it non projective sentence.

2. Long distance labeling . Example:-"RAm ney seb khaya or phir seeta ney aam khaya ,kyuki wah bahut bhooki thi"

3. Improving inter chunk labeling.

4. Mixed language parsing.

**4.2.6  Named Entity Recognizer(NER)**

It is also called as Entity Chunker/identifier/extractor.It is the process of identification of entities that are proper nouns and classifying them to their appropriate predefined tags like person-name, organization-name, time, monetary value etc.Gazetteer is a list which stores the named entities.The challenges for NER are mainly due to missing capitalization of noun words, ambiguity in proper and common-noun, spell variations in names, and free word order. For example in sentence "This article was written by Ram", the capital case of proper noun "Ram" helps to identify it as named entity. But the Hindi languages don't use any such capital cases.The words like "kamal" can be used as proper noun(lotus) and special noun (named entity- person name). As Hindi are free order languages, the position of word cant be used as indicator of the entity.

Saha et al [100] have suggested 2 methods for generating gazetteer for IL.First is based on transliteration of English gazetteer data to IL and second is domain specific method where seed context patterns regarding usage of NE are used to bootstrap similar patterns for other NE.For transliteration, as English has 26 and Hindi has 52 alphabets, so direct mapping is not possible.So the authors have suggested a new intermediate 34 characters alphabet to be used for transliteration.This approach gave proper transliteration for 91.59% words from total 1070 Hindi words.And using this intermediate form the gazetteer list can be prepared.However this method is applicable only if the parallel language NE list is available.For second method context pattern in window size of 3 was considered with the name of entity in middle.

As was observed in survey of previous 15 years conducted in 2007 by Nadeau [78], the methods used for NER had shifted from hand coded rules to machine learning based methods and gave better results.In 2008, a research [41]using hybrid approach of mixing CRF with rules got accuracy of only 50.06%.They used suffix like "baad" are used to identify the named entity in absence of capitalization feature, but failed to identify domain specific entities like "conditional random field". Saha et al[101] on using hybrid method of using MaxEntropy model with gazetter and context patterns to identify named entity, got accuracy of 81.52% on 4 NE classes. The researchers Ekbal et al [40]got results of 80.93% using CRF language dependent manually made gazetter and other language independent features, for Hindi NER on the dataset given in a contest.However, they mapped the confusing tags to common tag groups, and worked on only 4 categories from total 12 categories given in contest. In 2011, Srivastava et al[125] proved that if sufficient NE tags are not present in the tagged dataset, the simple rule based NER gives better result compared to statistical approaches of CRF and MaxEnt. In 2012, group of researchers Chopra et al[32] claimed the hybrid system of rules followed by HMM, gave results of 94.61% when testing for 9 self made tags.A different language independent rule based method[81] using matching transliterated entities from Hindi to English named entity database, based on their sound property was found to give 80.2% precision on testing for 3 tags in parallel corpus.In 2016 Chopra et al [33]experimented with HMM using tag set given by IIIT-H of 12 classes and got accuracy of 97.14% on testing with 105 tokens and training on 2343 tokens%.

---
[19]http://www.cs.cmu.edu/ ark/TurboParser/

A recent survey in 2019 for NER[105] in Hindi has found that hybrid system using rules with HMM is best till date.
As of now no special NER tools are released for open access for Hindi languages. Based on the survey following can be considered open research issues for NER task-

1. Domain specific tags, time , measurement tags handling.

2. Efficient gazetter generation and maintenance.

3. Mixed code NER tagging.

### 4.2.7 Word Sense Disambiguation (WSD)

It is the process which helps to find correct sense as per context for a multi-sense word.Example the sentence -"The bank was muddy" has ambiguous word "bank" .But given the context of "river" ,it can be fixed to "the shore of river". Methods used for WSD can be based on using hand coded rules which use the knowledge encoded in dictionary like wordnet/wikipedia/thesaurus etc. Another method for WSD is machine learning using naivebayes, SVM etc but it needs large sense annotated corpus.WSD can also be done in unsupervised way, based on the concept that sense of word is dependent on sense of neighbouring words.Such unsupervised methods use clustering or co-occurrence graphs.The knowledge based machine readable dictionary lookup methods are fundamentally overlap algorithms which suffer from overlap sparsity, as dictionary definitions are generally small in length.In 2012, Mishra et al [76]suggested a semi-supervised method wherein they selected few seed instances of ambiguous words and manually tagged the sense and based on this small set of training, and then classifier was trained on it.This method gave precision accuracy of 61.7% on 605 self made test instances of 40 words.In 2013, a knowledge based method using the semantic measure was attempted for WSD[115]. The measure was based on the length of paths between noun concepts in an "is-a" hierarchy, instead of relying on direct overlap the algorithm.The testing was done on 20 words in 710 instances to give 60.65% precision which was almost 15% more than using direct overlap measure.

In a survey done by Chandra et al in 2014[29], it was proved that the supervised machine learning approaches for WSD, were giving better results than all rest approaches.In same year of 2014, Singh et al [114]gave a WSD for Hindi using naive Bayes supervised algorithm which needed morphological analysis to transform words to base noun form prior to disambiguation.To train their system they had designed own corpus of 7506 instances of 60 words and got precision accuracy of 86.11% using window size of 5.The same researchers [113]tried another knowledge based method of WSD on same dataset.They used semantic relations available in Hindi wordnet, and found that the relation of hyponymy helps to improve the WSD precision by 9.86% and when all 5 semantic relations are used total 12.09% improvement found as compared to using direct Hindi word net(HWN) which gave only 50.6% precision accuracy. In 2009 [61], work by Khapra et al suggested cross linking of synsets across Marathi, Hindi and English languages.Next the wordnet parameters were projected from the resource rich language to resource poor language, to get the sense probabilities. They used the concept that the sense remains similar within a domain, irrespective of language.The achieved accuracy of Marathi WSD was 71.86% which was 14% better than using question language wordnet which did not have all sense tagged. In most latest research on Hindi WSD in 2019 by Sharma et al[104], 71% accuracy was reported using more enriched Hindi wordnet apply knowledge based algorithm where the context words in sentence are matched with the various words used in the examples and gloss of the ambiguous word.Testing was done on 3k words.

No publicly released tools are found for Hindi WSD.Thus it can be concluded the results are not very encouraging and lots of work needs to be done for WSD.

### 4.2.8 Anaphora Resolution(AR)

AR is the problem of resolving references in the discourse of a text. These items are usually pronouns like in English (he, she, they etc) and in Hindi-वह, खुद, मैं, तुम, जब –तब etc.The AR can be intrasentential or intersentential. Below are examples showing sentences with type of AR needed.

1. "Seeta ate banana.Later she drank water"-intersentential.

2. "Seeta ate banana and then she also drank water"-intrasentential.

3. "She who ate banana was Seeta"-intrasentential.

There are 3 types of pronouns references based on position of their referred items namely anaphoric (previous item referred ), cataphoric(later item referred) or exophore(outside of discourse item referred).In above examples 1 and 2 are anaphoric, 3 is cataphoric and the sentence like "He who is smart would succeed" is exophoric. Some of the challenges in anaphora resolution for IL are -

- Gender knowledge not useful Ex: pronoun उसे refers to both girl and boy.While in languages like English, His/her clearly indicates the gender of referred noun so resolution can be less difficult.

- Non availability of standard efficient tools.If AR has correctly tagged POS sentences, morphological analyzed words and parse structure, it would be easy to identify the pronoun and nouns.But these tools are still not efficiently developed for IL.

For Hindi, Agarwal et al[1] used rule based approach which needed words with grammatical attributes namely gender, count, type of noun, animate and extra information like name of river, location, revered person etc.The noun word which had maximum matching attributes with pronoun, was resolved as the connected noun.The testing was done on 200 sentences having pronouns, which were assumed to be referring to previous 2 sentences to max.The accuracy of 96% for simple case (where the needed noun was in previous sentence and it was the subject in first noun phrase) but only 80% for complex and compound sentences was found. Another rule based AR for dialogue system was given in 2015 by Mujadia et al[77].The system had accuracy of 75% for first person pronoun but only 35% for relative pronoun like जो (which), जसे(to which), etc.The testing was done on 25 dialogues with total 1986 sentences using nearly 3500 pronouns.The researchers had annotated data which used paninian dependency grammar based parser to get the needed syntactic semantic information of words, named entity information, animacy and subtopic boundary information . In recent rule based method given by Tewani et al[127] in 2020 for Hindi AR, grammatical knowledge about the animistic (living thing/non living thing) and number (singular/plural), was used to get 82% accuracy.The work had considered only two previous sentence for finding referred entity.The testing is done on 3 stories of maximum 22 lines.

### 4.2.9 Toolkit

A tool kit provide group of processing resources in a bundle of same package. The 2 prominent toolkits/libraries available are 1) iNLTK and (2)Indic NLP Library (2) StanfordNLP.

The iNLTK is Python library specially helpful to developer making applications in Indic languages.The toolkit uses language models trained on wikipedia articles[20].iNLTK provides text tokenization, language identification, extract embedding vectors etc. The IndicNLP libarary is for researchers working on NLP in Indic languages.It is available from IITB [21].It provides a Python based library of programs for text normalization, sentence splitter(simple rule based system), script Information(gets phonetic information of each symbol like is it vowel, nasal etc), tokenization(based on punctuation boundaries), word segmentation(gives component morphemes of inflected or derived word), script conversion(among IL which are having unicodes offset same from respective base), romanization (convert text to roman script text), indicization(convert roman script text to IL script text), transliteration (amongst IL) and translation(amongst Indian languages and between Indian and English language). The StanfordNLP[22] is also using neural language models in Python supporting tokenization,MWE identification, Lemmatization, POS tagging and also parser, but only for Hindi Language.

**Summary of Processing tools** The table 8 gives a concise statistical scenario of processing tools in Hindi languages as of now.

### 4.3 Enabling Resources

The resources which help in standardization of resources enabling consistent NLP system development are put under the "enabling resources" category.

---

[20]https://inltk.readthedocs.io/en/latest/
[21]http://anoopkunchukuttan.github.io/indic_nlp_library/
[22]https://nlp.stanford.edu/

Table 8: Summary:Processing tools

| Tool name | Language | Design approaches | | | Annotated corpus available | Tool released |
| --- | --- | --- | --- | --- | --- | --- |
| | | Symbolic | Statistical | Other | | |
| Tokenization : | H+E codemix | 78.58% | Sup-NB:84.5% | — | — | yes |
| Tokenization: | Hindi | — | — | Hybrid :Rule + sup- Dtree:98.6% | — | yes |
| MA-Stemmer | Hindi | 83% | Unsup-89.9% | — | — | no |
| MA-lemmatizer | Hindi | 89.08% | — | — | — | yes |
| Morph analysis | Hindi | 99% | Sup-svm:82.03% | yes | yes | — |
| POS tagger | Hindi | 87.55% | sup-maxent:94.38% | Connectionist-ANN:91.30% | yes | yes |
| Chunker | H+E codemix | — | Sup-RF:73.3% | — | — | — |
| | Hindi | — | Sup-hmm:91.70% | — | yes | yes |
| | H+E codemix | — | Sup-CRF:78% | — | — | — |
| Parser | Hindi | 87.6%LA | Sup-92.90%LA | Constraint based-76.90%LA | yes | — |
| WSD | Hindi | 71% | sup-NB:86.11% | — | — | — |
| AR | Hindi | 82% | — | — | yes | — |
| NER | Hindi | 80.2% | sup-hmm:97.14% | — | yes | — |

### 4.3.1 Scripts

For Hindi the "Devanagari" script is used. The script used by English is named Latin script or Roman script.

Unicode is an abstract mapping of characters from all of the world's scripts to unique numbers(code points) [23].In this mapping scheme, the range for Hindi it is 0900 to 097F. There are various encoding schemes like UTF8/16/32, ASCII etc to convert unicode to binary.The UTF-8 encoding mechanism is widely used for Indic Scripts.One benefit of UTF-8 is the backward compatibility to the ASCII encoding, which was used for Latin Script of English symbols earlier.

### 4.3.2 Tag sets

A tagset is a list of labels used to indicate the part of speech/Chunks in text corpora.Main tag sets available for Hindi are listed below -

- POS and/or chunk tagset
    - IIIT-Hyderabad/ILMT tagset for POS and chunk- [24] 21 categories of POS and [25] 8 categories for Chunk set.Flat tagsets based on Penn tagset of English language.Specifically made for MT application(Anncora).
    - LDC-IL tagset for POS- Managed by CIIL, Mysore[26].Hierarchical basic categories which can be extended to add more categories as per language need.Independent of any target application[14].Hindi has 13 basic categories.
    - Universal POS tagset-[91] Includes 22 languages(no IL). Only 12 tags.
    - BIS (Bureau of Indian standards) POS tagset-Last tagset released(in 2010).Hierarchical tagset with 2 levels. As per experimentation done by Nitish in 2014, this tagset gives best precision performance of 88.2%[82] amongst all tag sets for Indic NLP.

- Named entity tagset
  Given by Ministry of Communication and IT, GOI, developed at AU-KBC research center [27], Anna University, Chennai.It is hierarchical tagset of total 102 tags with 22 tags at first level.FIRE provides tagged data as per this tagset in its tracks related to NER.

**Summary : Language specific resources and Enabling resources** The table9 shows the size of language specific consolidated resources available as of now for Hindi language.

Table 9: Summary:Language tools

| Name of Resource | Language | Size |
| --- | --- | --- |
| Corpus-raw | Hindi (H) | 1M words |
| corpus-MA annotated | Hindi | 30k words |
| Corpus -treebank | Hindi | 3000 sentences |
| corpus-POS and chunked | Hindi | 36 k sentences |
| corpus-Coreferrence tagged | Hindi | 3600 sentences (dependency + constituency) |
| parallel corpus | H -E pair | approx. 57k sentences |
| word list | Hindi | 168k |
| dictionary | E-H | 88k words |
| named entity | Hindi | 2k words |
| MWE | Hindi | 15k words |
| wordnet | Hindi | 105423 words /40465 synsets |

## 5 Applications

There are many applications which access/use the data in natural languages to have some particular goal satisfied like translate data from one language or format to other, extract relevant information, summarize the information etc. Typically in Indian languages, it is found that majority of data is written using mix of scripts and languages. The applications using these setups are known as multi script or multi language/multilingual applications respectively.Further, if the input is strictly in one script and information stored uses another script, the application is known as cross-script.Now, the below sections would discuss the state of art in these applications for Hindi languages. Few well known applications are as shown in figure 7 and which would be now explored in detail in upcoming section.

---

[23]https://unicode.org/main.html
[24]http://ltrc.iiit.ac.in/tr031/posguidelines.pdf
[25]ltrc.iiit.ac.in/MachineTrans/research/tb/InstraChunkDependencyAnnotationGuidelines.pdf
[26]http://www.ldcil.org/standardsTextPOS.aspx
[27]http://au-kbc.org/nlp

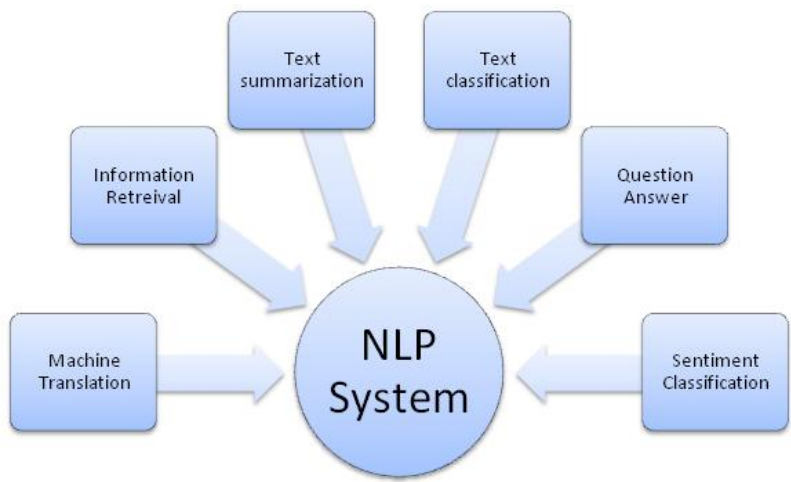

Figure 7: NLP Applications

## 5.1 Machine Translation(MT)

The goal of MT : "Output a most appropriate translated text in target language 'y' corresponding to input text given in 'X' language". There are mainly 2 broad approaches widely used for MT namely rule based (RBMT) and corpus/data based(CBMT) as shown in figure8.

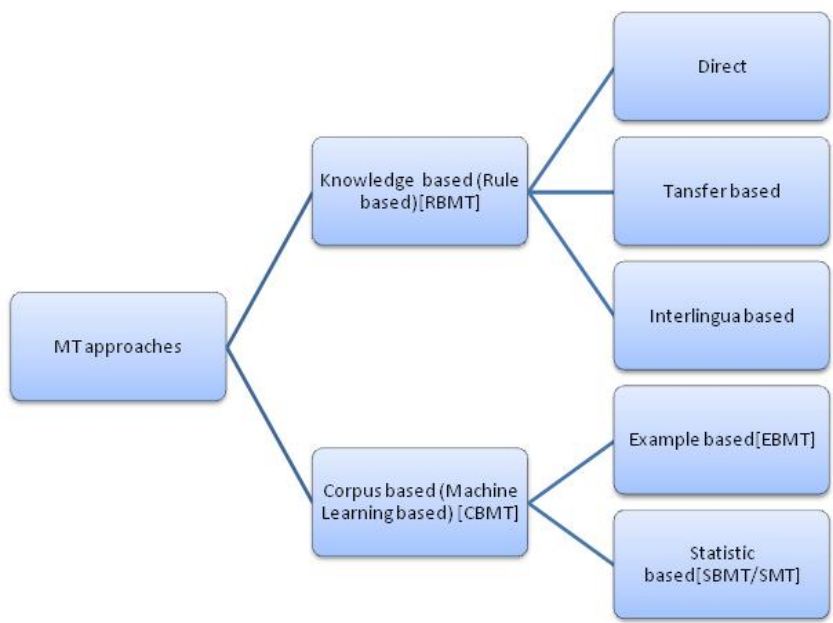

Figure 8: Machine Translation taxonomy

The CBMT can be further be classified into EBMT and SBMT or SMT. EBMT maintains all possible patterns/examples of parallel texts and fetches the appropriate text based on some similarity metrics.Such systems have high search time.The SMT systems don't need much linguistic background and is also mathematically sound process, but they would need huge parallel corpus. Based on the level of knowledge used to do MT, the RBMT systems can be described using the triangle as shown in figure 9[130]

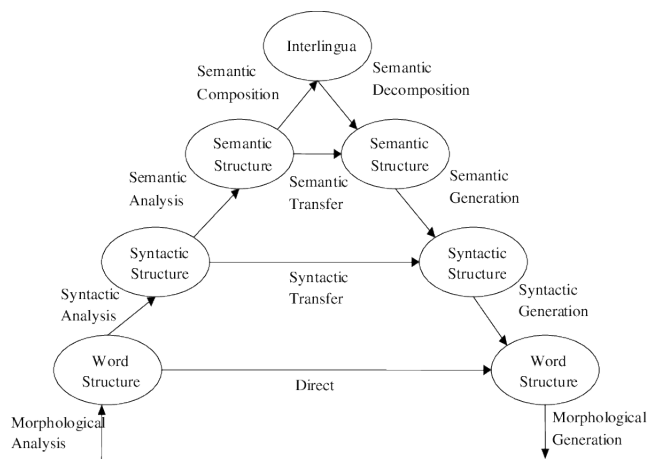

Figure 9: Rule based MT methodologies

In the direct RBMT, word to word translation is done using bilingual corpus lookup and it expects the languages to be similar.Another mechanism to translate is where the input text can be converted to some intermediate form(intform) which can in turn be used to generate the appropriate target text.Now if this intform uses syntactic structure like tree to transfer the data to target, it is called transfer based MT and if the intform uses a special intermediate notation which is independent of any language, then it is called interlingua based method.Thus, direct MT works on morphological level, transfer based MT works at syntax/semantic level While the interlingua MT works till pragmatic level.The major challenge with RBMT is the need of language expertise to design the rules.

Currently the noteworthy ongoing-projects in context of English-Hindi MT are as follows [79, 43]-

1. Rule based : Interlingua based - at IITB for E-H using UNL(Universal Networking Language) [37];

2. Rule based:Transfer based-matra, mantra and Anuvdaksh at CDAC for E->H and E->G.

3. Corpus based:EBMT-Anubharti at IITK for E->H ;Shiva-shakti at IIITH and IISC Bangalore, transfer rule based with EBMT for E->H;

4. Corpus based:SMT-phrase based at NewYork University for H->E; EILMT which does SMT with rules for E->H at CDAC; Sampark by IIITH, using SMT with transfer based method for MT amongst IL.

5. Hybrid: AnglaBharti at IITKanpur(IITK) for E<->H using PLIL(Psuedo lingua for IL) rule based with example base,with accuracy 90%[121].

**Other noteworthy developments in MT for Hindi** A large sized well designed corpus is released for Hindi by Bojar et al in 2014[26]. For Hindi-English codemixed data, heuristic rule based method to translate the mixed chunks into entire Hindi or English text was given by Sinha[120].This translation gives accuracy of nearly 90% and is added to the already existing MT system for Eng<->Hindi MT system (Anglabharti). The widely used Google translate was found to be having 9.5 BLEU score for English to Hindi (E->H) MT [2], however it is dated 2011 and no recent analysis was found.The IndicNLP library from IITB, provides phrase based SMT for E->H with 30.86, E->G of 21.33 and G<->H of 53.06 BLEU scores[66] as of 2014.

Singh et al proposed to apply postprocessing rules on the Hindi sentence received from E->H MT system.Their post processing method corrected 75% of sentences from tested 100 sentences,however they have not integrated to any MT to find overall improvement of MT system[116].

In 2017,Singh et al suggested to use recursive recurrent neural network model based deep learning to improve the MT for E-H [117], but implementation was not attempted.A method of interactive SMT two phase system is suggested for Hindi by Jain in 2020.In first step the most probable translated texts which match the source text and the result of SMT are given to user and then in second step the user selects and modifies if needed the translation. This method was found to improve the speed and productivity of 5 human translators by nearly 50% [53].

### 5.1.1 Transliterator

A transliterator application helps to changes the letters from one alphabet or language into the corresponding, similar-sounding characters of another alphabet or language.For example input Hindi word हंस in Devanagari scrip can be transliterated as "haans" in roman script for English. This module can help the machine translator, for OOV words like for names of person, places etc. The current scenario in MT for Hindi is shown in table 10.The table indicates the active organizations working for inter language translations amongst Hindi(H) and English(E) and the reported efficiency (if any).Note the score indicates BLEU unless otherwise stated.

Table 10: Machine Translation:Current status

| Languages Source->Target | RBMT Direct | RBMT Transfer based | RBMT Interlingua based | CBPT EBMT | CBPT SMT | Hybrid |
|---|---|---|---|---|---|---|
| E->H | --- | IITB (30.86); IIITH ; CDAC | IITB (using UNL) ; | IIITH and IISC | CDAC, IITH | IIITK (RBMT, PLIL EBMT ) " |
| H->E | --- | CDAC; IITH | --- | IITK | NYU | |
| E->G | --- | CDAC, IITB (21.33) | --- | --- | --- | --- |
| H->G | 88 (BLEU) [89] | --- | --- | --- | --- | --- |
| G->H | --- | CDAC, IITB (53.06) | --- | --- | --- | --- |
| G->H (transliteration) | 93.09% [5] | --- | --- | --- | --- | --- |
| Hinglish->Hindi or English | --- | --- | --- | 90% accuracy [120] | --- | --- |

## 5.2 Information Retrieval(IR)

IR system's main task is retrieval of all relevant information from provided sources like corpus with text, images,videos etc. We here would be concentrating on IR systems using natural language in written form.As shown in figure 10 such NLP IR system, can be classified into 4 types based on the text features.If the corpus is stored in well formed data structures like tables, it is called structured IR else unstructured. If the data to be searched is looked up from web, it is called search engine IR and if the query text and document text are in different languages it becomes a cross language IR.

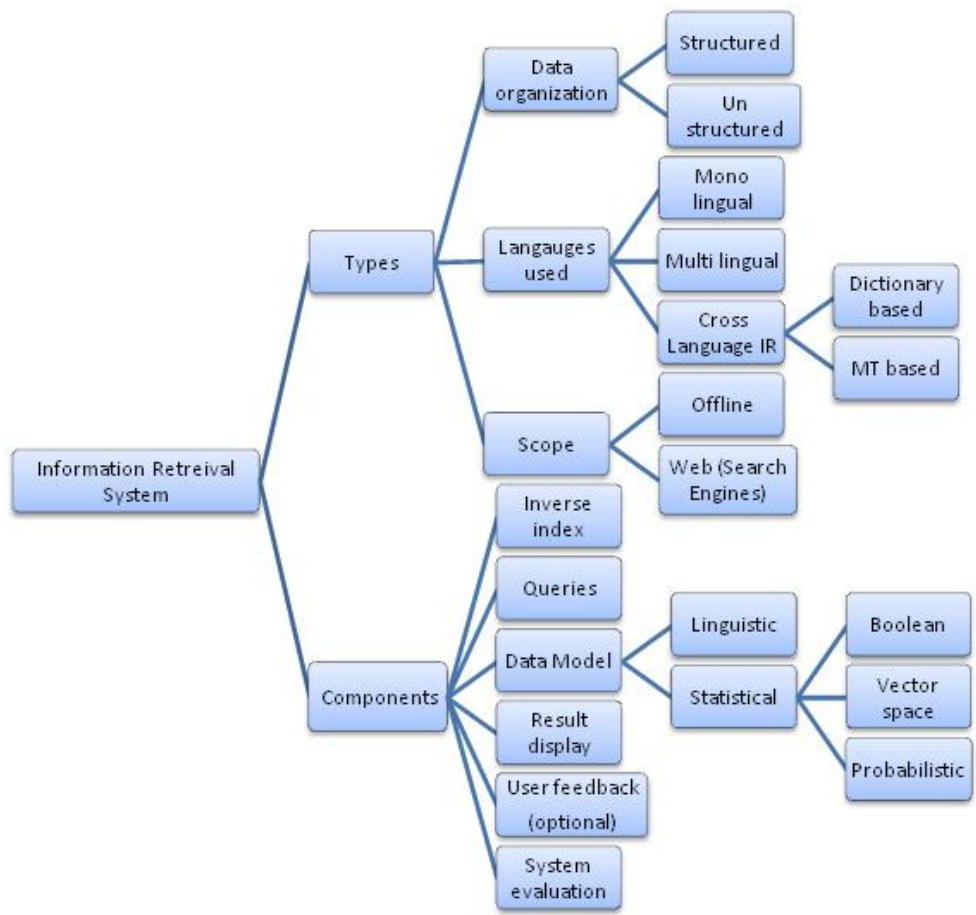

Figure 10: NLP based IR Taxonomy

Further the typical NLP IR systems which are generally unstructured type rather than structured type, consists of following major components- [74]

1. Inverse indexing - Mapping from terms(keywords) to documents in corpus.

2. Queries or topics with details of information needs- To identify what is to be searched from the corpus.The query needs to be framed in a way so as to find relevant information. Example- If information need is "Get name of person who killed Ravan",and query is framed as "Who killed Ravan", then the text "Though it was clear who killed Ravan,nothing can be said about his character", is excellent match for query but still is not "relevant".

3. Data representation models- Documents/queries should be represented in most efficient manner which can help in efficient retrieval. There are two widely used approaches to design these models: 1) Statistical approach. 2)Linguistic approach.Generally the mix of the both approaches is used in IR system. The statistical methods emphasize on "bag-of-words" concept while the linguistic approach considers the meaning of the words to find relevance i.e "bag-of-senses".The linguistic approach tries to find the meaning of documents by using analysis of words, structure of documents, context based word meaning disambiguation etc. [129]. The major models used for statistical approach are:

    - Boolean model- Operators like "and", "or" used to form query . Terms presence or absence marked in boolean. Uses similarity measure like Jaccard to retrieve documents for given query.
    - Vector space model- Use term frequency, Inverse document frequency (IDF) and other values of document to represent it as vector.Measures like cosine similarity can be used to establish the relevance of documents with reference to query.
    - Probabilistic model- Using classification or clustering methods to identify probability of relevance of documents with context to given query.

4. Display results- Rank the retrieved results and display in appropriate sequence.

5. Take feedback from user [optional]-it would help to redefine the query to improve the system. If no feedback taken, it is called adhoc IR system.

6. Evaluation- The relevant document and query pairs are provided for evaluation of IR system.The measures to quantify the judgements are -recall,precision,MAP(mean average precision),F-score etc.In IR, recall is considered more important then precision. Few methods to improve recall are (1)Do query expansion using synonyms (2) use latent semantic indexing for better match of documents with queries (3) Relevance feedback from user to improve query. It is found that MAP gives best judgement in presence of multiple queries.The relevance of retrieved documents is considered more important than speed of retrieval [95].

There are many international research groups/conferences which aid for development of IR systems, like TREC(Text Retrieval conference) [28] and CLEF (Conference and Labs of the Evaluation Forum, [29]. CLEF was mainly working for multi language IR, which includes Hindi. FIRE, formed in 2008, is the forum of Indian retrieval evaluation designed on similar grounds for promoting IR development in Indian languages.

For Hindi and other IL, a search engine named Webkhoj was designed by Pingali and group in 2006.The authors ran the crawler for 6 months daily and retrieved half million pages and analyzed the results.Hindi language websites were maximum but not many sites were using standard Unicode encoding and hence were not indexed and search able by the regular search engines.The authors hence suggested to convert the non-standard encoding of IL sites to standard encoding[92]. Currently Google provides the search facility in Hindi, but it still has huge scope of improvement as it fails to handle ambiguous word, fails to find the synonymous word based expansion, does not have proper context based results etc.Example:-For the query नारी को सोना पसंद है : naari koo soona pasand hai ,the Google returned pages with reference to sleep ,whereas the meaning of "soona" here was "gold". Another search engine for Hindi is released by TDIL named Sandhan, but it is needs lots of updations.

For CLIR as different languages are used in query and documents language translation is needed in either or both.There are 3 major approaches for translation of document and query - (1) using dictionary (2) by MT tool (3) parallel corpora.Generally query are translated by dictionary or parallel corpora while documents use MT systems for translation [70]. By query expansion, MAP of 0.428 was achieved for CLIR where documents were in Hindi and query in English by Larkey et al in 2003[68].The English query was expanded based on the top5 terms in top 10 English documents retrieved using initial query. The expanded query is then translated to Hindi. The researches also found that normalization,stopword removal and transliteration of OOV words, improve the CLIR. FIRE had provided a data of 1.9 GB of Hindi documents for the IR task. A report from FIRE organizers in 2013 confirms that monolingual Hindi IR has MAP of 0.44 but when query was in English and corpus in Hindi it drops to 0.3771[71]. Raza in a recent survey related to query expansion approaches, found domain specific ontologies are more preferable for expanding query in contrast to general ontologies [98]. In 2019,another approach by Jain et al, was taken for query expansion using Hindi wordnet. They gave different weights to relations in wordnet, based on their semantic importance. And for query expansion the higher weighted relations were given preference. For example hypernymy-hyponymy is given more weightage then meronymy -holonymy relation and hence for car, the "vehicle" relation is considered more important then "wheel". They tested method on 95k Hindi documents collected from FIRE. It is found to improve the Hindi IR system as compared to system using direct wordnet relations expansion [52]. However they have not used MAP to give the exact results obtained. Query expansion using word embedding was done by Bhattacharya et al in 2018[24].The Google translate,dictionary translations and cluster based on word embeddings were used for query expansion. The Hindi to English CLIR gave precision of 0.225 when tested on FIRE data. In 2016, Sharma used wikipedia to translate the Hindi query(n-grams) to English.If the n-gram query was found on Wiki,its counterpart English phrase was used to form new query in English.The CLIR had documents in English taken from FIRE.The method gives only 0.26 MAP, as not many proper interwiki links for Hindi terms were found[106].In 2011 , Sivakumar proved that [122]on using latent semantic indexing with singular value decomposition and TFIDF normalization,the performance of system increased three times as compared to when direct matching.The testing was done on parallel corpus of total 360 single line documents in Hindi and English. For CLIR testing, the English query was fired on the Hindi documents and it was expected the parallel document is retrieved correctly.Similarly when Hindi query is used, search is done in English corpus.FIRE had organized a mixed script, cross language IR track from 2013 till 2016 where IL and English languages were considered.They have presented a comprehensive summary of all observations[12], some major are as follows.(1)If script had used mix of words from languages only average of 0.201 F-score was obtained whereas if words were entirely English words then F-score was 0.807 and if entirely Indian languages then 0.574 on average.(2) Machine learning algorithms have got better accuracy then using dictionary and rule based approaches.Especially naivebayes was found better then SVM, CRF, logistic regression algorithms.

For developing efficient IL websearch engine (sandhan), efforts were made by priyatam and group, to make a domain focused webcrawler[93]. They were successful to get recall of 0.75 ,for language (Hindi) and domain specific focused crawling on "health" domain.

---

[28]https://trec.nist.gov/
[29]http://www.clef-initiative.eu/

Recent survey has found that there is still need to develop IR systems with better precision and recall using machine learning for specific domains like agriculture, medical etc.[123]. Another line of research can be to find user's search history to find his interests and preferences, use multiple ontologies, etc., to do query expansion. The major factors effecting efficiency of IR is query quality, quantity of data, languages used and dependent resources available for making the system.

### 5.2.1 Question Answering (QA)

QA is special type of IR.Here the data retrieved (answer) can be word, sentence, paragraph or an entire document, which answers the query precisely.Whereas the retrieved data in IR are typically entire documents from where seeker has to skim to find what she/he exactly needs. The figure 11 shows the major phases and types of such systems.

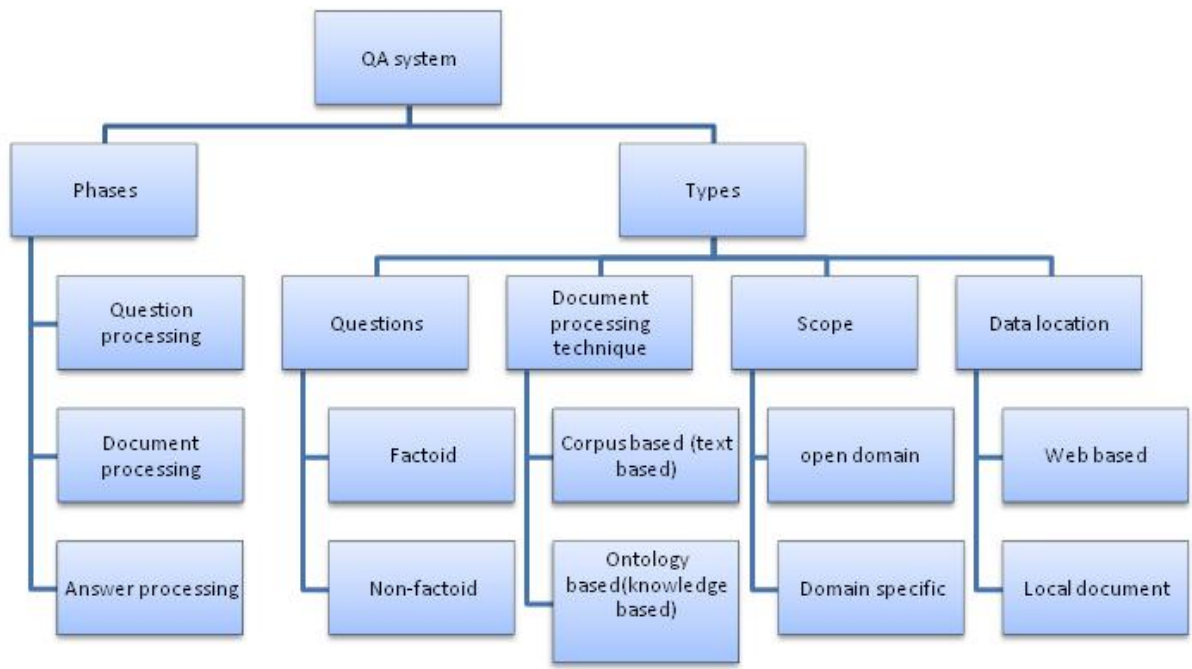

Figure 11: Question-Answer systems :Taxonomy

As shown in figure,generally the main tasks of these systems are -

- Question processing - process the question, find keywords, find answer types based on class identified of question, expand and reformulate query.If the question is in cross language, translation also might be needed. The questions can be fact based or non-fact based.The not-factual are more difficult to answer.

- Data processing - Retrieve exact answer from given data-sources which might be online or offline.The data might be in form of text documents or ontology based knowledge structures. If the corpus of documents is used as source for answer finding, it would need the IR techniques for finding relevant answers.And if the ontology is to be used to find answer, the system refers to knowledge based structures to extract answer. The knowledge based data structures use special languages/frameworks like OWL[30],RDF[31] etc.and special query language like SPARQL[32] is needed to access these ontological structures.Generally the corpus based processing is preferred for open domain QA systems whereas for domain specific QA systems, the ontological method is suitable.

- Answer processing- Validate and rank the answers and display.

The restricted domain question answering system (RDQA), are found to give higher accuracy as compared to open domain QA, which would need WSD. Rule based QA uses heuristic rules to identify classes of question and answers expected like "where" refers to location, "when" refers to time based answer etc. Web based QA finds answers from web.The questions can be of various complexity levels and various types like interrogative, affirmative etc. The system might not need full parser, but only shallow parser using MA, POS tagger and chunker would be sufficient [48]. In 2005, it was tried to make Hindi domain specific QA for agriculture and science domain but as no Hindi wordnet was available, the authors Kumar et al, had made own synonym list, MA, stemmer, stop word removal and available IR tool for English was modified to handle Hindi. The system was reported to give average 75% accuracy on 30 questions under each domain[65]. In FIRE 2015, QA track was started but it did not include Hindi [12].

---

[30]https://www.w3.org/OWL/
[31]https://www.w3.org/TR/rdf11concepts/
[32]https://www.w3.org/TR/rdf-sparql-query/

As per a survey in 2018 done on Hindi QA by ray et al[97], there was lots of scope for improvement in these systems.Few open research options were, development of better resources and tools,development of non-factoid questions-answers corpus,develop standard evaluations and test beds, develop knowledge based system working at discourse level, etc. Attempt was made by Anuranjana et al in 2019, to frame questions from sentences based on dependency parse structure for Hindi [8].They were able to get 100 questions from 30 sentences but they have worked only for simple questions generation.

## 5.3 Text Classification(TC)

The task of text classification is to assign classes(labels) to an unlabelled document.Based on the various purpose/target of classification it can be named as sentiment/opinion/emotion classification, document classification, email classification.It can also be used by web crawlers to classify the web-pages for focused IR and for spam detection.The text classification can be domain specific like for business, finance, industry, internet, patents etc.It can be used for language identification and classification in applications using multi language or codemixed languages. There are various types of TC whose names are self explanatory - binary(two target class), multi-class, multi-label(more then one class to each text), hierarchical(classes in hierarchy). A study suggests that India is in top 10 countries in research publications related to this field [75].They also found that from 2 major types of text representation models namely vector space and graph, the former is used more often.The various majority of features used are n-gram words or words related to aspect/emotion or embedded vectors, named entities. The feature weighing methods used are mainly TFIDF, Binary, BM25(best matching) and recent trends is using the deep features, genetic algorithms, semantic based weighing. For training the classifier model various methods like supervised, semi-supervised, active (training algorithm also adds labels), transfer(improve current model based on learning from related task), ensemble(more than one methods combined to train) and multi-view(use data from different view feature set).For evaluation most prevalent is recall, precision, F-score and average.Some other measures are also used like error rate, CPU training and test time, memory needed by model. For feature reduction and selection, methods like SVD, PCA and LDA are applied. Recent attempt for Hindi TC using CNN based deep learning model was giving 92.8% accuracy [58].However as the researchers did not have sufficient data in Hindi, they had used the Google translator to convert the data available under TREC and Stanford sentiment data sets. And after converting to Devanagari script, the data sets were used.The FIRE is also providing interesting tracks related to this task like Indian native language identification of user from the text written by user in English [6]. Another track was related to hate and offensive content classification from Hindi text on social media [73].

### 5.3.1 Sentiment Classification (SC)

SC is a type of TC where the target is to find the opinion/sentiment/emotion in a piece of text with respect to some topic.The basic SC classes are positive, negative or neutral.SC can be done by using resources like sentiwordnet, lexicon which has POS tags with sentiment values or other such lexical resources.However these resources are priorly assigned values so might need revisions to suit current situations and needs.Another method is machine learning based which is more robust but it needs large amount of sentiment labeled corpus.Other related subtopics in SC which are researched in IL context are sarcasm/ controversy/ rumour/ fake data detection [39].The issues and more details of sarcasm classification are discussed in [57]. The first reported work in Hindi SA was reported to have 78.14% accuracy and had used annotated self made corpus with SVM [56].In 2014 the task to build the sentiment lexicon for Hindi from related English words in wordnet by graph propagation was done and for Hindi 3640 were included in the developed lexical sentiment resources [31]. Major issue in this lexicon is insufficiency of negative sentiment words. One study was done in mixed script Indic sentences. The approach where language specific sentiwordnet was used was found to be 8% better than using MT to convert all sentences to English and then doing SA on those English sentences [21].

## 5.4 Text Summarization(TS)

It is the task of condensing and providing the compact version of the given original text without changing the underlying meaning of text.If the sentences of original text are re-framed in the summary text then it is known as abstractive TS, while extractive TS fetches most important sentences from original text to make summary. TS can categorized based on its focus as query based or topic based summarization.The main steps in such systems are preprocessing, feature extraction and then sentence scoring , ranking and finally summary generation. The last two phases are language independent.The feature extraction can be done based on language dependent linguistic resources like NER, wordnet, word embeddings, sentiwordnet etc.The TS can also use the language independent features like word counts, sentence score etc.Evaluation of system can be done manually or using automatic measures like precision, recall and F-score.The ROUGE(Recall oriented understanding for Gisting evaluation) is widely used algorithm for finding the overlap between human generated summary and system generated summary.

Few work has been reported in Hindi TS.One method is by Gulati et al [47] where they have done extraction based TS using statistic and linguistic features. They reported 73% precision.Kumar had given a method for single document summarization using semantic similarity to get recall of 69% at 60% compression but at higher compression of 80% the recall reduced to 45% [64].The method of using Hindi wordnet for POS tagging to facilitate feature extraction followed by using genetic algorithm for selection and ranking is suggested by Thaokar and group[128] but they had not disclosed the results.

Patel et al [85] had given language independent approach for extractive TS. They suggested to use only structural and statistical features and no semantic features so the method can be generalized for any language TS. They proved that their system was giving 80% representative summaries for any language. One toolkit is also released by Jhaveri [55] which provides cross language TS. It is tested with the source text in English and summary generated in Hindi, with 4.13 and 7.59 ROUGE F-scores respectively. They had taken the data from DUC (Document understanding conferences) and manually converted the summaries from English to IL for given source documents. DUC is run by NIST-National Institute of Standards and technology, USA and its main goal is to encourage research in TS by providing data sets, tools and other software for TS. The toolkit has modules for sentence simplification, inbuilt graph based model, machine translation, bootstrap code, developing new model and evaluation module using ROUGE implementation. Recent review by Baruah in 2019 suggests that current TS are designed to work at syntactic level in IL[13]. Another research group found that English has better tools and hence it gets 0.41 F-score as compared to Hindi with 0.37 [132]. They pointed that there are issues like non standard stopword list,stemmer, NER, underdevelopment wordnet for IL, which affect the TS.

# 6 Analysis and Findings

Following are few major observations of the survey in NLP for Hindi languages-

- Many of developed tools/resources are not released in open domain in Hindi.So standardization of testing not possible.

- The language resource of wordnet(WN) still has scope of improvement.AS of today, English wordnet has 77194 more synsets then Hindi WN .Also there are lots of standard thesaurus and annotated corpus available for English.

- The MT was mostly from English to Hindi, but from Hindi to English not much effort found.

- For IR application, cross language IR improvement is needed for Hindi to English IR systems.

- The TS applications in Hindi are made mostly using extractive methods, abstractive techniques need to explored.

- There are many open issues like sarcasm detection, irony detection, humour identification, stance detection etc. which are yet not fully explored in IL under sentiment classification.

# 7 Conclusion

In the present survey, the information about the organizations working in the Hindi NLP systems, were presented. Also a taxonomy of various major tools developed in Hindi NLP systems as of date, was given. And later a taxonomic survey of the tools was used to give the current state-of-art and subsequently to present the research gaps. This survey might help a researcher and industrialist to identify the feasibility, scope and limitations of NLP systems using Hindi languages. The survey paper also gave overview of the scenario in Hindi based NLP systems for few trending applications like Machine translation,Information retrieval,text classification and sentiment analysis. In future, the study would be extended to more application domains like Information extraction, recommendation system, topic linking, argumentation mining, text prediction, dialogue system, argumentation mining, recommendation system etc.

[93] Pattisapu Nikhil Priyatam, Srikanth Reddy Vaddepally, and Vasudeva Varma. Domain specific search in indian languages. In *Proceedings of the first workshop on Information and knowledge management for developing region*, pages 23–30, 2012.

[94] Avinesh PVS and G Karthik. Part-of-speech tagging and chunking using conditional random fields and transformation based learning. *Shallow Parsing for South Asian Languages*, 21:21–24, 2007.

[95] Prabhakar Raghavan. Information retrieval algorithms: A survey. In *Proceedings of the eighth annual ACM-SIAM symposium on Discrete algorithms*, pages 11–18, 1997.

[96] Ananthakrishnan Ramanathan and Durgesh D Rao. A lightweight stemmer for hindi. In *the Proceedings of EACL*, 2003.

[97] Santosh Kumar Ray, Amir Ahmad, and Khaled Shaalan. A review of the state of the art in hindi question answering systems. In *Intelligent Natural Language Processing: Trends and Applications*, pages 265–292. Springer, 2018.

[98] Muhammad Ahsan Raza, Rahmah Mokhtar, Noraziah Ahmad, Maruf Pasha, and Urooj Pasha. A taxonomy and survey of semantic approaches for query expansion. *IEEE Access*, 7:17823–17833, 2019.

[99] Dwijen Rudrapal, Anupam Jamatia, Kunal Chakma, Amitava Das, and Björn Gambäck. Sentence boundary detection for social media text. 2015.

[100] Sujan Kumar Saha, Sudeshna Sarkar, and Pabitra Mitra. Gazetteer preparation for named entity recognition in indian languages. In *Proceedings of the 6th Workshop on Asian Language Resources*, 2008.

[101] Sujan Kumar Saha, Sudeshna Sarkar, and Pabitra Mitra. A hybrid feature set based maximum entropy hindi named entity recognition. In *Proceedings of the Third International Joint Conference on Natural Language Processing: Volume-I*, 2008.

[102] Dikshan N Shah and Harshad Bhadka. Paradigm-based morphological analyzer for the gujarati language. In *Intelligent Communication, Control and Devices*, pages 469–481. Springer, 2020.

[103] Arnav Sharma, Sakshi Gupta, Raveesh Motlani, Piyush Bansal, Manish Srivastava, Radhika Mamidi, and Dipti M Sharma. Shallow parsing pipeline for hindi-english code-mixed social media text. *arXiv preprint arXiv:1604.03136*, 2016.

[104] P Sharma and N Joshi. Knowledge-based method for word sense disambiguation by using hindi wordnet. *Engineering, Technology & Applied Science Research*, 9(2):3985–3989, 2019.

[105] Richa Sharma, Sudha Morwal, and Basant Agarwal. Named entity recognition for hindi language: A survey. *Journal of Discrete Mathematical Sciences and Cryptography*, 22(4):569–580, 2019.

[106] Vijay Kumar Sharma and Namita Mittal. Exploiting wikipedia api for hindi-english cross-language information retrieval. *Procedia Computer Science*, 89:434–440, 2016.

[107] Jikitsha Sheth and Bankim Patel. Dhiya: A stemmer for morphological level analysis of gujarati language. In *2014 International Conference on Issues and Challenges in Intelligent Computing Techniques (ICICT)*, pages 151–154. IEEE, 2014.

[108] Manish Shrivastava, Nitin Agrawal, Bibhuti Mohapatra, Smriti Singh, and Pushpak Bhattacharya. Morphology based natural language processing tools for indian languages. In *Proceedings of the 4th Annual Inter Research Institute Student Seminar in Computer Science, IIT, Kanpur, India, April*, 2005.

[109] Manish Shrivastava and Pushpak Bhattacharyya. Hindi pos tagger using naive stemming: harnessing morphological information without extensive linguistic knowledge. In *International Conference on NLP (ICON08), Pune, India*, 2008.

[110] Ajeet Singh, Bhanu Pratap Singh, Ankit Kumar Poddar, and Abhishek Singh. Sentence boundary detection for hindi–english social media text. In *Recent Findings in Intelligent Computing Techniques*, pages 207–215. Springer, 2018.

[111] Akshay Singh, Sushma Bendre, and Rajeev Sangal. Hmm based chunker for hindi. In *Companion Volume to the Proceedings of Conference including Posters/Demos and tutorial abstracts*, 2005.
32